\documentclass[runningheads]{llncs}
\usepackage[T1]{fontenc}
\usepackage{graphicx,verbatim}
\usepackage{amsmath,amssymb}
\usepackage{booktabs}
\usepackage{multirow}
\usepackage{xcolor}
\usepackage{subcaption}
\usepackage[hidelinks]{hyperref}
\usepackage{microtype}
\usepackage{amsmath}
\usepackage{amssymb}
\usepackage{amsfonts}

\newcommand{\ours}{MolDA}

\begin{document}

\title{MolDA: Molecular Understanding and Generation via Large Language Diffusion Model}
\titlerunning{MolDA: Molecular Understanding and Generation via LLM Diffusion}
\author{Seohyeon Shin$^{1,*}$ \and
HanJun Choi$^{1,*}$ \and
Jun-Hyung Park$^{2}$ \and
Hong Kook Kim$^{1,\dagger}$ \and
Mansu Kim$^{1,\dagger}$ 
}

\authorrunning{S. Shin and H. Choi et al.}

\institute{Gwangju Institute of Science and Technology, Gwangju, South Korea\\
\email{\{ann4622.sin, gkswns1290\}@gmail.com, \{hongkook, mansu.kim\}@gist.ac.kr}
\and
Hankuk University of Foreign Studies, Seoul, South Korea\\
\email{jhp@hufs.ac.kr}}

\maketitle
\renewcommand{\thefootnote}{}
\footnotetext{$^{*}$Equal contribution. $^{\dagger}$Corresponding authors.}
\begin{abstract}
Large Language Models (LLMs) have significantly advanced molecular discovery, but existing multimodal molecular architectures fundamentally rely on autoregressive (AR) backbones. This strict left-to-right inductive bias is sub-optimal for generating chemically valid molecules, as it struggles to account for non-local global constraints (e.g., ring closures) and often accumulates structural errors during sequential generation. To address these limitations, we propose MolDA (\textbf{Mol}ecular language model with masked \textbf{D}iffusion with m\textbf{A}sking), a novel multimodal framework that replaces the conventional AR backbone with a discrete Large Language Diffusion Model.
MolDA extracts comprehensive structural representations using a hybrid graph encoder, which captures both local and global topologies, and aligns them into the language token space via a Q-Former. Furthermore, we mathematically reformulate Molecular Structure Preference Optimization specifically for the masked diffusion.  Through bidirectional iterative denoising, MolDA ensures global structural coherence, chemical validity, and robust reasoning across molecule generation, captioning, and property prediction.

\end{abstract}

\section{Introduction}

Large Language Models (LLMs) have emerged as a strong foundation for building general-purpose intelligence, motivating their extension to the molecular domain for drug discovery and materials science~\cite{mol_instructions}. Early molecular LLMs primarily represented molecules as one-dimensional strings (e.g., SMILES or SELFIES) and applied autoregressive (AR) generation to tasks like property prediction and molecule captioning~\cite{molt5,chemdfm,llasmol}. However, purely sequential representations obscure the native topological relationships that govern chemical interactions. To preserve this crucial structural information, recent multimodal architectures integrate graph neural network (GNN) encoders, aligning graph features with the LLM embedding space via cross-modal projectors~\cite{liu2023molca,park2024llamo}.

Despite these architectural advances, current multimodal molecular LLMs still fundamentally rely on AR backbones~\cite{fang2024moltc,liu2023molca}. This strict left-to-right inductive bias is sub-optimal for molecular generation, where chemical validity heavily depends on non-local, global constraints such as ring closures and valence satisfaction~\cite{selfies}. Because AR models lack access to future context during generation, early local decisions can invisibly accumulate errors, leading to invalid global structures~\cite{msr}. Recently, discrete diffusion language  models (DLM) have emerged as a compelling alternative, framing text generation as an iterative denoising process from a fully corrupted sequence~\cite{llada,sahoo2024mdlm}. By enabling non-AR, bidirectional generation, diffusion models allow for the continuous revision of tokens based on global consistency. While diffusion has shown promise in 3D conformation generation~\cite{diffSBDD,geodiff}, its application to holistic, language-based molecular understanding remains largely underexplored~\cite{chemllmreview,tgm_dlm}.

To address this gap, we propose \textbf{\ours{}}, a multimodal framework replacing the conventional AR backbone with a DLM, LLaDA-8B-Instruct~\cite{llada}. Beyond this architectural shift, \ours{} introduces two key methodological innovations. First, to mitigate modality imbalance (i.e., graph bypass), we mathematically reformulate Molecular Structure Preference Optimization (MolPO) by redefining implicit rewards based on the masked diffusion log-likelihood. Second, we design task-specific sampling strategies—full-sequence pure diffusion for molecule generation and block diffusion with low-confidence remasking for text—to better capture non-local atomic constraints such as ring closures. This allows \ours{} to attend to global structural coherence during generation, while leveraging the diffusion backbone for molecular understanding tasks.

\section{Method}
\begin{figure}[!b]
\centering
\includegraphics[width=0.9\textwidth]{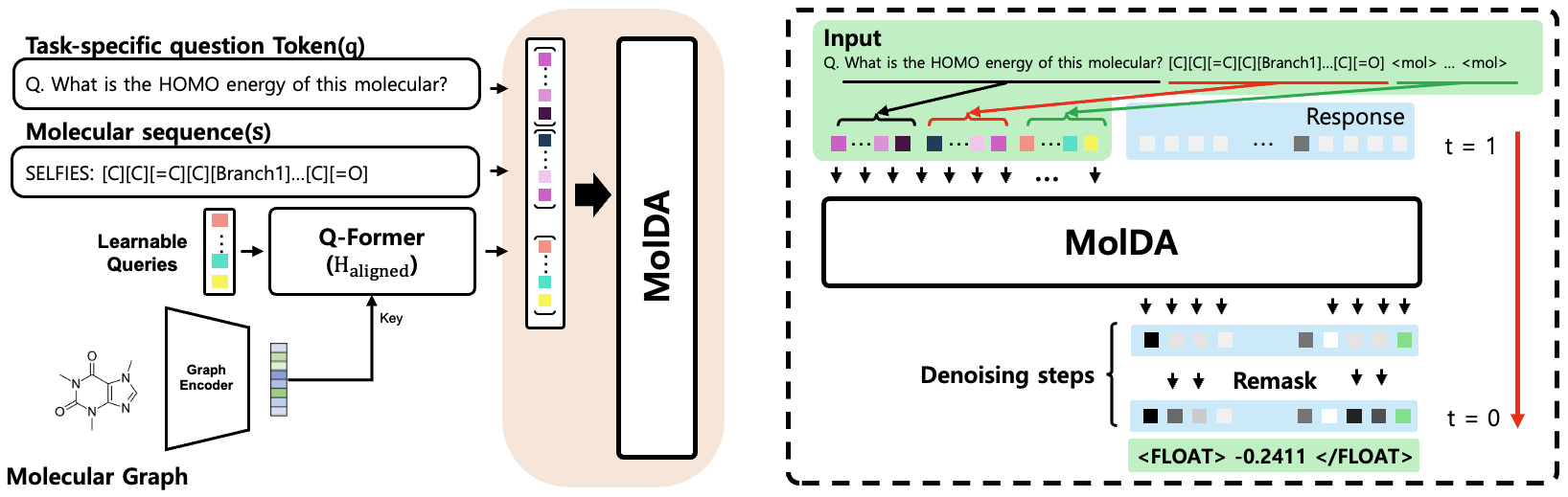}
\caption{\ours{} architecture overview. A hybrid graph encoder (GINE + TokenGT) produces structural representations, the Q-Former maps them into the language model token space, and LLaDA generates the response through iterative denoising with low-confidence remasking.}
\label{fig:architecture}
\end{figure}

The overall workflow of the proposed model (i.e., MolDA) is illustrated in Fig.~\ref{fig:architecture}. Given a 2D molecular graph paired with a natural language instruction which contain question with a SELFIES~\cite{selfies},  MolDA  handles diverse tasks including property prediction, reaction prediction, retrosynthesis, molecule captioning, and text-guided molecule generation.

\subsection{Architecture of MolDA}
\paragraph{\textbf{Hybrid Graph Encoder}} To capture both local substructures and global topology, we employ a hybrid encoder integrating GINE~\cite{gine} and TokenGT~\cite{tokengt}. Let $\mathcal{G} = (\mathcal{V}, \mathcal{E})$ denote a molecular graph. The encoder consist of two parallel branches. First, the local branch utilizes GINE to encode neighborhood interactions, yielding graph-level $\mathbf{h}_g^{\text{GINE}} \in \mathbb{R}^{1 \times d_g}$ and node-level $\mathbf{H}_v^{\text{GINE}} \in \mathbb{R}^{|\mathcal{V}| \times d_g}$ embeddings: $\mathbf{h}_g^{\text{GINE}}, \mathbf{H}_v^{\text{GINE}} = f_{\text{GINE}}(\mathcal{G})$. Simultaneously, the global branch employs TokenGT to capture long-range dependencies by using nodes and edges as tokens, producing graph-level $\mathbf{h}_g^{\text{GT}}$, node-level $\mathbf{H}_v^{\text{GT}}$, and edge-level $\mathbf{H}_e^{\text{GT}}$ embeddings: $\mathbf{h}_g^{\text{GT}}, \mathbf{H}_v^{\text{GT}}, \mathbf{H}_e^{\text{GT}} = f_{\text{TokenGT}}(\mathcal{G})$. Here, $d_g$ denotes the embedding dimension (set to 1024). Finally, we concatenate outputs from both branches to form the unified representation $\mathbf{H}_{\text{hybrid}} \in \mathbb{R}^{(2|\mathcal{V}| + |\mathcal{E}| + 2) \times d_g}$:
\begin{equation}
    \mathbf{H}_{\text{hybrid}} = [ \mathbf{h}_g^{\text{GINE}} \,;\, \mathbf{H}_v^{\text{GINE}} \,;\, \mathbf{h}_g^{\text{GT}} \,;\, \mathbf{H}_v^{\text{GT}} \,;\, \mathbf{H}_e^{\text{GT}} ].
\end{equation}
Subsequently, we utilize this hybrid graph embedding as the input for the Q-Former to align representation.
\vspace{-1mm}
\paragraph{\textbf{Cross-Modal Alignment Projector}} 
To bridge the representation gap between the graph encoder and the LLM backbone, we employ a Q-Former as a cross-modal projector. To efficiently transform the variable-length graph representation $\mathbf{H}_{\text{hybrid}}$ into a fixed sequence of aligned features, we utilize $N_q$ learnable query tokens $\mathbf{Q} \in \mathbb{R}^{N_q \times d}$ (set to 32) to extract and compress structural features. The queries interact with the graph features via multi-head cross-attention, where $\mathbf{Q}$ acts as queries and $\mathbf{H}_{\text{hybrid}}$ serves as keys and values. The aligned molecular embedding $\mathbf{H}_{\text{aligned}} \in \mathbb{R}^{N_q \times d}$ is computed as:
\begin{equation}
    \mathbf{H}_{\text{aligned}} = \text{Softmax}\left(\frac{\mathbf{Q} (\mathbf{H}_{\text{hybrid}} \mathbf{W}_K)^T}{\sqrt{d_k}}\right) (\mathbf{H}_{\text{hybrid}} \mathbf{W}_V),
\end{equation}
where $\mathbf{W}_K, \mathbf{W}_V$ are learnable projection matrices. These aligned tokens $\mathbf{H}_{\text{aligned}}$ are subsequently concatenated with task instructions and SELFIES tokens, forming the conditional input for the diffusion backbone.
\paragraph{\textbf{Language model Backbone}}
To overcome the sequential bias of AR framework, MolDA employs \textit{LLaDA-8B-Instruct}~\cite{llada}, to generate text response through discrete diffusion process that iteratively refines the entire sequence. Unlike AR models that generate tokens $x_t$ conditioned solely on history $x_{<t}$, LLaDA captures the joint probability of the entire sequence, enabling structural coherence.

The training of MolDA involves a forward diffusion and a reverse denoising process. Let $\mathbf{q}$ denote the task-specific question tokens (e.g., ``Please provide a detailed description of the molecular structure''), $\mathbf{s}$ denote the SELFIES sequence of the input molecule, and $\mathbf{x}_0$ denote the clean target response sequence. Depending on the downstream task, $\mathbf{x}_0$ takes various forms, such as a molecule caption, a target SELFIES string, a numerical value, or a boolean label.

In the forward process, we gradually mask tokens in $\mathbf{x}_0$ based on a time step $t \in (0, 1]$. Specifically, each token is independently replaced by a special \texttt{[MASK]} token with probability $t$. Formally, the transition is defined as: $q(x_t^i | x_0^i) = (1 - t) \cdot \mathbf{1}_{x_t^i = x_0^i} + t \cdot \mathbf{1}_{x_t^i = \texttt{[MASK]}}$, where $q(x_t^i | x_0^i)$ denotes the conditional transition probability of the $i$-th token, and $\mathbf{1}$ is the indicator function. 
Eventually, at $t=1$, the sequence becomes entirely masked. In the reverse process, we recover $p_\theta(\mathbf{x}_0)$ starting from $\mathbf{x}_1$ (fully masked) and iteratively denoising over $N$ steps. Finally, MolDA takes $\mathbf{x}_t$, $\mathbf{q}$, $\mathbf{s}$, and $\mathbf{H}_{\text{aligned}}$ as inputs and predicts all masked tokens simultaneously. The training objective minimizes the variational lower bound, simplifying to the following weighted negative log-likelihood:
\begin{equation}
    \mathcal{L}_{\text{DLM}} = -\mathbb{E}_{t, \mathbf{x}_0} \left[ \frac{1}{t} \sum_{i=1}^{L} \mathbf{1}_{x_t^i = \texttt{[MASK]}} \log p_\theta(x_0^i \mid \mathbf{x}_t, \mathbf{q}, \mathbf{s}, \mathbf{H}_{\text{aligned}}) \right],
\end{equation}
where $L$ denotes the sequence length of $\mathbf{x}_0$. Through this optimization, the model learns to infer missing tokens conditioned on the visible sequence and the global graph topology, ultimately achieving structural coherence.

\subsection{Training process}
\paragraph{\textbf{Domain-Adaptive Tokenization.}} 
Before describing the multi-stage training process, we define our molecular representation. We represent molecules using SELFIES~\cite{krenn2020selfies}, as it provides built-in syntactic constraints. Because the original LLaDA tokenizer lacks dedicated tokens for SELFIES (e.g., \texttt{[C]}, \texttt{[=N]}, \texttt{[Ring1]}), we expand its vocabulary by adding 2,944 SELFIES-specific tokens. The embeddings for these new tokens are initialized by sampling from a normal distribution matching the mean and standard deviation of the pre-trained embeddings. 
\paragraph{\textbf{Hybrid Graph Encoder and DLM pretraining}}
To learn comprehensive molecular representations, we initially pretrain the hybrid graph encoder. Specifically, the graph encoder is optimized via two auxiliary tasks: functional group prediction and SELFIES reconstruction~\cite{mol_llm}. First, to capture local chemical properties, a three-layer MLP $f_\theta$ predicts the presence of functional groups from the aligned features $\mathbf{H}_{\text{aligned}}$. This is optimized using a binary cross-entropy loss:
\begin{equation}
    \mathcal{L}_{\text{func}} = -\sum_{k=1}^{K} \left[ y^{(k)} \log \hat{y}^{(k)} + (1 - y^{(k)}) \log(1 - \hat{y}^{(k)}) \right],
\end{equation}

\noindent where $\hat{y}^{(k)} = f_\theta(\mathbf{H}_{\text{aligned}})^{(k)}$ denotes the predicted probability for the $k$-th functional group, $y^{(k)} \in \{0, 1\}$ is the ground-truth binary label, and $K$ is the total number of functional groups.

Second, to encode global structural semantics, we reconstruct the corresponding SELFIES sequence utilizing an AR GPT-2 decoder $\pi^{\text{GPT-2}}_\phi$. Here, aligned features $\mathbf{H}_{\text{aligned}}$ serve as the context to predict each SELFIES token $s_i$:
\begin{equation}
    \mathcal{L}_{\text{recon}} = -\sum_{i=1}^{L_s} \log \pi^{\text{GPT-2}}_\phi(s_i \mid \mathbf{H}_{\text{aligned}}, \mathbf{s}_{<i}),
\end{equation}

\noindent where $\mathbf{s}_{<i}$ denotes the preceding tokens of the SELFIES sequence, and $L_s$ is its sequence length. The overall pretraining objective for the graph encoder is formulated as $\mathcal{L}_{\text{GNN}} = \mathcal{L}_{\text{func}} + \mathcal{L}_{\text{recon}}$.

Prior to multimodal integration, we perform supervised fine-tuning (SFT) on the DLM backbone using our text-only instruction-tuning dataset. This step injects molecule-specific prior knowledge into the language model and significantly reduces the computational overhead during the subsequent multimodal training phase. Building upon the discrete diffusion framework described previously, we employ a text-only masked diffusion objective~\cite{llada}. For a given target textual sequence $\mathbf{x}_0$ of length $L$ and a uniformly sampled masking ratio $t \in (0, 1]$, we obtain the partially masked sequence $\mathbf{x}_t$. The backbone is optimized to reconstruct the original tokens strictly at the masked positions:
\begin{equation}\label{eq:loss_sft}
    \mathcal{L}_{\text{SFT}} = -\mathbb{E}_{t, \mathbf{x}_0} \left[ \frac{1}{t} \sum_{i=1}^{L} \mathbf{1}_{x_t^i = \texttt{[MASK]}} \log p_\theta(x_0^i \mid \mathbf{x}_t, \mathbf{q}, \mathbf{s}) \right],
\end{equation}

\noindent where the normalization factor $\frac{1}{t}$ balances the expected number of masked tokens across different time steps. This objective relies solely on the internal textual contexts ($\mathbf{q}$ and $\mathbf{s}$) without external graph conditioning.

\paragraph{\textbf{Cross-Modal Alignment via Q-Former}}
During the cross-modal alignment stage, we freeze the weights of both the pre-trained hybrid graph encoder and the DLM backbone, exclusively updating the parameters of the Q-Former projector. This targeted updating strategy prevents catastrophic forgetting of the pre-trained unimodal knowledge while efficiently establishing cross-modal connections. The Q-Former is trained for one epoch by optimizing $\mathcal{L}_{\text{SFT}}$, as formulated in Eq.~\ref{eq:loss_sft}.

\paragraph{\textbf{Molecular Structure Preference Optimization}}
In standard SFT of multimodal molecular models, the language model backbone often suffers from modality imbalance. Specifically, the model heavily relies on the 1D textual sequence while largely ignoring the explicit topological features provided by the 2D molecular graph. To encourage the model to actively utilize structural information, we adopt Molecular Structure Preference Optimization (MolPO)~\cite{mol_llm}, which optimizes the representation preference between an original (chosen) graph $\mathcal{G}_w$ and a structurally perturbed (rejected) graph $\mathcal{G}_\ell$.

To generate the rejected graph $\mathcal{G}_\ell$, we strictly follow the perturbation strategy proposed in Mol-LLM~\cite{mol_llm}. Specifically, rather than relying on complex, task-specific heuristics, we adopt their MACCS keys-based functional group modification. By randomly replacing inherent substructures within the original graph $\mathcal{G}_w$, this method efficiently disrupts the alignment between the graph topology and the target response. This provides a generalized and computationally lightweight mechanism for preference learning across diverse downstream tasks.

While the original MolPO framework was designed for AR next-token prediction, we mathematically adapt it to our discrete diffusion framework. Let $\mathbf{H}_{\text{aligned}}^w$ and $\mathbf{H}_{\text{aligned}}^\ell$ denote the cross-modal embeddings obtained by processing $\mathcal{G}_w$ and $\mathcal{G}_\ell$ through the hybrid graph encoder and the Q-Former, respectively. We formulate the rewards $r_{w} = \frac{\beta}{N_{\text{mask}}} \sum_{i=1}^{L} \mathbf{1}_{x_t^i = \texttt{[MASK]}} \log p_\theta(x_0^i \mid \mathbf{H}_{\text{aligned}}^w, \mathbf{x}_t, \mathbf{q}, \mathbf{s})$ and $r_{\ell} = \frac{\beta}{N_{\text{mask}}} \sum_{i=1}^{L} \mathbf{1}_{x_t^i = \texttt{[MASK]}} \log p_\theta(x_0^i \mid \mathbf{H}_{\text{aligned}}^\ell, \mathbf{x}_t, \mathbf{q}, \mathbf{s}),$ where $\beta$ controls the reward scaling, and $N_{\text{mask}}$ denotes the number of \texttt{[MASK]} tokens in the partially masked sequence $\mathbf{x}_t$ as the average log-likelihoods over the masked tokens. The final MolPO objective optimizes the preference margin between the chosen and rejected graphs using a clipped log-sigmoid loss:
\begin{equation}
    \mathcal{L}_{\text{MolPO}} = -\mathbb{E}_{\mathbf{x}_0, \mathbf{H}_{\text{aligned}}^w, \mathbf{H}_{\text{aligned}}^\ell} \left[ \log \sigma \Big( \beta \cdot \left( \min(r_w - r_\ell, \lambda_{\text{clip}} |r_w|) - \gamma_m \right)\Big) \right],
\end{equation}

\noindent where $\sigma(\cdot)$ is the sigmoid function, $\lambda_{\text{clip}}$ prevents excessive penalization of the rejected reward by clipping the margin, and $\gamma_m$ serves as a task-adaptive target reward margin for the $m$-th molecular task. The overall multimodal training objective for MolDA is formulated as $\mathcal{L}_{\text{Total}} = \mathcal{L}_{\text{SFT}} + c \mathcal{L}_{\text{MolPO}}$, where $c$ is a balancing constant.

\subsection{Inference}\label{sec:inference}
MolDA generates tokens via an iterative discrete diffusion process. Starting from a fully masked sequence of length $L$, the model progressively unmasks tokens over $N$ denoising steps by predicting the distribution over all masked positions simultaneously:
\begin{equation}
    \hat{\mathbf{x}}_0 = \arg\max p_\theta(\mathbf{x}_0 \mid \mathbf{x}_t, \mathbf{q}, \mathbf{s}, \mathbf{H}_{\text{aligned}}).
\end{equation}

To efficiently recover sequences, we adopt task-adaptive sampling strategies~\cite{llada}. For standard natural language tasks (e.g., molecule captioning), we employ \textit{block diffusion} with \textit{low-confidence remasking}. This block-by-block generation selectively retains high-confidence predictions while remasking uncertain ones, leveraging bidirectional context to resolve local ambiguities. Conversely, for molecule generation tasks (e.g., SELFIES), we utilize a \textit{full-sequence pure diffusion} approach. Because molecular validity heavily relies on non-local atomic constraints like ring closures, block-wise inductive biases can disrupt structural coherence. By simultaneously predicting and remasking low-confidence tokens across the entire sequence, the model continuously attends to global molecular topology. Finally, across all strategies, the output is obtained by truncating the refined sequence at the first predicted \texttt{[EOS]} token.

\section{Experiments and results}
\subsection{Experimental setup}
\paragraph{\textbf{Data description}}
We adopt the same four instruction-tuning datasets as Mol-LLM~\cite{mol_llm}, SMolInstruct~\cite{llasmol}, Mol-Instructions~\cite{mol_instructions}, ChEBI-20~\cite{chebi}, and PubChem comprising approximately 3.3M instances across eight tasks, with all sequences capped at 512 tokens (response $\leq$ 256).

\paragraph{\textbf{Implementation Details.}}
MolDA uses LLaDA-8B-Instruct~\cite{llada} ($\sim$8B params) as the backbone, a hybrid graph encoder (GINE $L{=}5$ + TokenGT, 224M params, $d_g{=}1024$), and a Q-Former with $N_q{=}32$ queries.
Training proceeds in three stages: Stage~1 pretrains the GNN (GINE 45 epochs, TokenGT 49 epochs, lr 1e-4) and the LLM via LoRA (15 epochs, lr 2.5e-4); Stage~2 trains the Q-Former only (1 epoch, lr 2.5e-5); Stage~3 jointly updates all components with MolPO (1 epoch, lr 4e-5).
All experiments use 8$\times$A100 40GB GPUs with bf16 mixed precision, and inference uses $T{=}64$ denoising steps.
We compare against six 7B--8B AR baselines: Mol-LLM~\cite{mol_llm}, ChemDFM~\cite{chemdfm}, LlaSMol~\cite{llasmol}, Galactica~\cite{galactica}, MolT5~\cite{molt5}, and 3D-MoLM~\cite{3d_molm}.


\begin{table}[!b]
\caption{Molecular understanding results on ChEBI-20 (Generation, Captioning) and MoleculeNet (Property Prediction). Best results are \textbf{bolded}, second best are \underline{underlined}.}
\label{tab:understanding}
\centering\footnotesize
\begin{tabular}{l cc cc cc cc}
\toprule
& \multicolumn{2}{c}{Generation} & \multicolumn{2}{c}{Captioning} & \multicolumn{2}{c}{Regression} & \multicolumn{2}{c}{Classification} \\
\cmidrule(lr){2-3} \cmidrule(lr){4-5} \cmidrule(lr){6-7} \cmidrule(lr){8-9}
Model
& Exact$\uparrow$ & MACCS$\uparrow$
& R-1$\uparrow$ & METEOR$\uparrow$
& LogD$\downarrow$ & HOMO$\downarrow$
& HIV$\uparrow$ & SIDER$\uparrow$ \\
\midrule
MolT5-Large
& .331 & .868
& \underline{.539} & \textbf{.480}
& - & -
& - & - \\
Mol-LLM
& \underline{.415} & \underline{.873}
& \textbf{.570} & \underline{.471}
& \textbf{0.981} & \textbf{.004}
& \textbf{.774} & \underline{.743} \\
Galactica
& .000 & .178
& .105 & .065
& 2.534 & .230
& .550 & .533 \\
LlaSMol
& .253 & .827
& .494 & .426
& \underline{1.582} & .982
& .685 & .622 \\
ChemDFM
& \textbf{.421} & \textbf{.891}
& .377 & .301
& 5.886 & .183
& .551 & .540 \\
3D-MoLM
& - & -
& .222 & .227
& 3.891 & .031
& .502 & .552 \\
\midrule
\ours{}
& .068 & .589
& .265 & .239
& 1.923 & \underline{.008}
& \underline{.761} & \textbf{.846} \\
\bottomrule
\end{tabular}
\end{table}

\subsection{Molecular Understanding.}
As shown in Table~\ref{tab:understanding}, \ours{} performs relatively poorly on ChEBI-20 generation and captioning compared to autoregressive generalist models, but achieves comparatively strong results on several property prediction benchmarks. In particular, \ours{} attains the highest SIDER AUROC of 0.846 and reasonably strong HIV and HOMO scores, although LogD and most regression metrics are still dominated by Mol-LLM. Overall, this suggests that the diffusion backbone is more beneficial for structure- and property-centric tasks than for pure text generation.

\subsection{Reaction Prediction.}
As shown in Table~\ref{tab:reaction}, \ours{} achieves the second-highest Exact Match across all three reaction prediction tasks on Mol-Instructions, and also attains the second-best MACCS scores for forward synthesis and reagent prediction. Most other generalist baselines obtain near-zero Exact Match, especially on forward synthesis.

\begin{table}[t]
\caption{Reaction prediction results on Mol-Instructions. Best results are \textbf{bolded}, second best are \underline{underlined}.}
\label{tab:reaction}
\centering\footnotesize
\setlength{\tabcolsep}{3pt}
\begin{tabular}{l cc cc cc}
\toprule
& \multicolumn{2}{c}{Forward Synthesis} & \multicolumn{2}{c}{Retrosynthesis} & \multicolumn{2}{c}{Reagent Prediction} \\
\cmidrule(lr){2-3} \cmidrule(lr){4-5} \cmidrule(lr){6-7}
Model
& Exact$\uparrow$ & MACCS$\uparrow$
& Exact$\uparrow$ & MACCS$\uparrow$
& Exact$\uparrow$ & MACCS$\uparrow$ \\
\midrule
Mol-LLM
& \textbf{.904} & \textbf{.985}
& \textbf{.512} & \textbf{.887}
& \textbf{.134} & \textbf{.535} \\
Galactica
& .000 & .215
& .000 & .283
& .000 & .134 \\
LlaSMol
& .038 & .676
& .026 & .650
& .000 & .200 \\
ChemDFM
& .302 & .808
& .080 & .769
& .000 & .229 \\
3D-MoLM
& .000 & .639
& .000 & \underline{.810}
& .000 & .175 \\
\midrule
\ours{}
& \underline{.662} & \underline{.907}
& \underline{.236} & .791
& \underline{.027} & \underline{.312} \\
\bottomrule
\end{tabular}
\end{table}

\subsection{Effect of Denoising Steps.}


We analyze the effect of the number of denoising steps $T$ on reaction prediction using the Stage~1 version of \ours{} (semi-AR decoding before applying MolPO). As shown in Table~\ref{tab:reaction_steps}, increasing $T$ from 32 to 64 improves both Exact Match and MACCS across all three tasks, while further increasing $T$ to 128 does not lead to consistent additional gains. Given the roughly linear increase in inference time with $T$, we use the more efficient setting with $T{=}64$ for all main MolDA results.

\begin{table}[t]
    \centering

    \centering
    \caption{Effect of denoising steps $T$.} 
    \label{tab:reaction_steps}
    \footnotesize 
    \begin{tabular}{c cc cc cc}
    \toprule
    & \multicolumn{2}{c}{Forward} & \multicolumn{2}{c}{Retro.}  \\ 
    \cmidrule(lr){2-3} \cmidrule(lr){4-5} 
    $T$ & Exact$\uparrow$ & MACCS$\uparrow$ & Exact$\uparrow$ & MACCS$\uparrow$  \\ 
    \midrule
    32  & 0.648 & 0.916 & 0.304 & 0.808 \\
    64  & 0.736 & 0.939 & 0.312 & 0.835  \\
    128 & 0.760 & 0.943 & 0.258 & 0.820  \\
    \bottomrule
    \end{tabular}


\end{table}

\section{Conclusion}
We proposed MolDA, a multimodal framework that replaces standard AR backbones with a DLM. By generating molecular sequences through iterative bidirectional denoising, MolDA addresses the error accumulation and structural constraint violations inherent to unidirectional decoding. To address modality imbalance, we reformulated MolPO for the masked diffusion objective, enforcing active utilization of 2D graph inputs. Empirical results demonstrate that MolDA achieves the best SIDER AUROC and competitive scores on several property prediction benchmarks, and achieves highly competitive accuracy in reaction prediction tasks. While AR models maintain an advantage in fluent text generation, this work  demonstrates that discrete diffusion is a viable backbone for multimodal molecular modeling. 
\begin{credits}
\subsubsection{\ackname}
This work was supported by the National Research Foundation of Korea (NRF) grant funded by the Korea government (MSIT) (RS-2024-00411137).
\end{credits}
\clearpage
\bibliographystyle{splncs04}

\bibliography{reference}

\end{document}